\documentclass[twoside,11pt]{article}
\usepackage{etoolbox}
\newtoggle{jair}
\togglefalse{jair}
\iftoggle{jair}{
\usepackage{jair}
\usepackage{theapa}
\usepackage{rawfonts}
}{
\usepackage[margin=2cm]{geometry}
}
\usepackage{url}
\usepackage[T1]{fontenc}
\usepackage[hidelinks=true]{hyperref}
\usepackage{appendix}
\usepackage{natbib}
\usepackage[utf8x]{inputenc}
\usepackage{booktabs}
\usepackage[disable]{todonotes}

\usepackage{microtype}
\usepackage{multirow}
\usepackage{placeins}
\usepackage{longtable}
\usepackage{enumitem}
\usepackage{amsmath}
\usepackage{cleveref}
\usepackage{wrapfig}
\usepackage{tipa}
\DeclareMathOperator*{\mean}{mean}
\iftoggle{jair}{
  \jairheading{70}{2021}{1-15}{10/2020}{1/2021}
\ShortHeadings{Visually grounded models of spoken language}{Chrupała}
\firstpageno{1}

\begin{document}

\title{Visually grounded models of spoken language:\\
  A survey of datasets, architectures and evaluation
  techniques}

\author{\name Grzegorz Chrupała \email grzegorz@chrupala.me\\
  \addr Department of Cognitive Science and Artificial Intelligence\\
  Tilburg University\\
  The Netherlands}
}{
  \begin{document}

\title{Visually grounded models of spoken language:\\
  A survey of datasets, architectures and evaluation
  techniques}

\author{ Grzegorz Chrupała\\ \url{grzegorz@chrupala.me}\\
  Department of Cognitive Science and Artificial Intelligence\\
  Tilburg University\\
  The Netherlands}
\date{}
}

\maketitle
\begin{abstract}
  \noindent
  This survey provides an overview of the evolution of visually
  grounded models of spoken language over the last 20 years.  Such
  models are inspired by the observation that when children pick up a
  language, they rely on a wide range of indirect and noisy clues,
  crucially including signals from the visual modality co-occurring
  with spoken utterances.  Several fields have made important
  contributions to this approach to modeling or mimicking the process
  of learning language: Machine Learning, Natural Language and Speech
  Processing, Computer Vision and Cognitive Science. The current paper
  brings together these contributions in order to provide a useful
  introduction and overview for practitioners in all these areas.  We
  discuss the central research questions addressed, the timeline of
  developments, and the datasets which enabled much of this work. We
  then summarize the main modeling architectures and offer an
  exhaustive overview of the evaluation metrics and analysis
  techniques.
\end{abstract}
\section{Introduction}
\label{sec:intro}

The endeavor of developing systems for modeling or simulating the
human language faculty has long been pursued by several disjoint
research communities. The speech community focuses largely on the task of
transcribing speech signals into written text, or extracting specific
pieces of information from spoken utterances. The field of Natural
Language Processing (NLP)
studies a range of problems involved in understanding and producing
language, but almost exclusively in its written form. Computer vision
researchers focus on vision and language tasks such as automated
captioning and visual question-answering. Meanwhile, cognitive scientists interested in language
develop models of language acquisition and processing within the
broader purview of understanding the cognitive makeup of our species
rather than with a view to practical applications.

The aim of the current survey is to review a family of approaches to
the computational study of language which largely transcend these
disciplinary divisions: more specifically, our focus is on visually
grounded models of spoken language. Such models are inspired by the
observation that when children pick up a language, they rely on a wide
range of indirect and noisy clues, including information from perceptual
modalities co-occurring with spoken utterances, most crucially the
visual modality.

Other sources of information such as phoneme and word co-occurrence
statistics, speaker intentions as inferred from gaze and gestures or
active learning are undoubtably important for children too, and
some have been modeled computationally.  In the interest of
tractability this survey zooms in on grounding spoken language in the
visual modality.

Many relevant studies have appeared during the last two decades, and
especially in the last five years, and as can be expected for a
research problem cross-cutting disciplinary boundaries, are spread
over publication venues focusing on Cognitive Science, Speech, Machine
Learning and NLP: here we bring them together in a single place in
order to provide a useful introduction and overview for practitioners
in all these areas.  

Most recent dcomputational approaches to modeling
spoken language have relied on artificial neural networks also know as deep learning.
A central concern regarding these approaches has been
understanding the nature and localization of representations which
they learn and we therefore also discuss in some detail the different
analytical techniques proposed for this purpose, as well as the main
findings resulting from their application.

\subsection{What this survey is and is not about.}
\label{sec:about}
In this survey we will focus on models of {\it spoken} language which
are visually grounded, that is models which process the audio signal
represented via low-level features such as the waveform, spectrogram
or mel-frequency cepstral coefficients (MFCC). Models which process 
written text are outside the scope of our interest here, whereas those
that rely on phonemic transcriptions will only be discussed briefly as
part of the background.

By {\it visual grounding} what is meant here are visual
representations of scenes associated with spoken utterances,
for example images that the utterances describe, or videos the
utterances occur in.\footnote{Most of work to date has focused on
  static images; the few studies involving video are discussed in
  \cref{sec:video}.}
There is a parallel strand of research on
exploiting visual information depicting the speaker and especially
their lips in addition to the audio channel for automatic speech
recognition (ASR): this endeavor is outside the scope of this survey.

\subsection{Central questions}
As mentioned above, visually grounded models of spoken language have
been of interest to several disciplines and as such the questions
these lines of research have been interested in answering have also been
quite diverse. From the point of view of cognitive science the main
concern is whether these models can help us understand the constraints
and the mechanisms in play for human children when they learn to
understand language: do they develop representations of phonemes? How
do they segment speech signal into word- or morpheme-like segments?
How do they match word forms with visual concepts and how are these
visual concepts formed themselves?

From the perspective of engineering-focused fields (ML, Speech and
NLP) the questions have often centered more on technical aspects such
as: what datasets are necessary to learn to match spoken utterances to
images? To what extent can this be done in an end-to-end fashion?  How
can performance be improved via engineering the neural architecture,
learning objectives or other aspects of training? In addition, the use
of large and opaque neural architectures has prompted researchers in
these fields to ask questions such as: what kind of information is
encoded in neural models, and how accessible it is in different
components? What are effective techniques for analyzing and
interpreting these representations, and according to which metrics can
they be evaluated? Is there is a correspondence between learned
representations and the core constructs hypothesized within phonology,
lexicon, syntax and semantics?  In the rest of the paper we will
explore answers to all these questions and some more.

\subsection{Two waves}
The work on visually grounded models of spoken language naturally
divides into two temporal clusters, or waves: (i) the period starting
in 1999 with Deb Roy's PhD dissertation \citep{roy1999learning} and
lasting until around 2005, and after a ten year break, (ii) the second
wave starting with the creation of the Flickr Audio Caption Corpus in
2015 \citep{harwath2015deep} and continuing until now. The current
survey is organized around this two-wave form. We will start by
reviewing some early efforts in \cref{sec:early}. We will then discuss
the importance of datasets in driving new developments in
\cref{sec:data}, and follow these developments in
\cref{sec:arch,sec:video},
and the variations in the set-up of the task and
applications in \cref{sec:variations}. Finally we will survey
approaches to evaluation of the systems and the analysis of learned
representations in \cref{sec:eval}.

\section{Early efforts}
\label{sec:early}

\paragraph{CELL}
Perhaps the first serious computational implementation of learning
spoken language via visual grounding was described by
\citet{roy1999learning}, and further developed in
\citet{roypentland2002learning} as the CELL (Cross-channel Early
Lexical Learning) model. The model originated in
\citet{roy1999learning}, and there were a number of subsequent
versions and tweaks of CELL
\citep[e.g.][]{roy2003grounded,mukherjee2003visual,roy2005towards};
here we discuss the canonical 2002 version.

The goal of the model was to perform three tasks: (i) segment speech
at word boundaries, (ii) form visual categories, and (iii) associate
words with these categories. The model partially relies on data to
learn these tasks: the data consists of speech recordings elicited
from participants while playing with pre-verbal infants using 42
objects from seven classes ({\it balls, toy dogs, shoes, keys, toy
  horses, toy cars, toy trucks}). These spoken utterances were paired
with different views of the objects used at the time of each
utterance. The amount of spoken language is described as approximately
1,300 utterances of around 5 words, for each of 6 participants. Around
8\% of words in the data were visually groundable.

The model includes a hard-coded visual system which extracts object
representations for camera images: these representations are shape
histograms designed to be invariant to position, scale and in-plane
rotation. Object color, size or texture are not captured. The
representations of spoken utterances consists of arrays of phoneme
probabilities extracted by a Recurrent Neural Network (RNN) model pre-trained
on the TIMIT dataset of phonetically transcribed speech
\citep{1993STIN...9327403G}.

The core part of the architecture is data-driven and consists of two
components. The short-term memory store buffers recent utterance-image
pairs; then utterances are segmented into subsequences (i.e.\ word
hypotheses) at phoneme boundaries located by the application of the
Viterbi algorithm to phoneme probabilities given by the RNN, and
subject to the constraint that a subsequence should contain at least
one vowel. Word hypotheses that occur more than once within the
buffer in a similar visual context are paired with co-occurring shape
representations and sent over to the long-term memory store. This
second component applies a filter to hypothesized associations,
keeping the ones which recur reliably (as measured by mutual
information) and discarding the rest.  The performance of the model is
quantified with respect to segmentation accuracy (do segment
boundaries correspond to word boundaries), word discovery (is the
segment a single English word?)  and semantic accuracy (is the word
matched with the correct shape?)  and compared to the baseline with a
similar architecture, but which ignores the visual modality: on all
measures CELL consistently outperforms the baseline by a substantial
margin.\footnote{According to \citet{roypentland2002learning},
  semantic accuracy can be measured for the baseline because the
  visual prototype was carried through from input to output and ``this
  model assumes that when a speech segment is selected as a prototype
  for a lexical candidate, the best choice of its meaning is whatever
  context co-occurred with the speech prototype.''}

The experiments with the CELL architecture demonstrated that learning
word meanings visually grounded spoken utterances was feasible. As
could be expected given the technology of the day, the approach did
have several important limitations: most importantly, the experiments
used small-scale data (the final lexicon evaluated included less than
20 learned form-meaning mappings) and the data for the visual modality
was rather artificial. Regarding modeling, the audio features were
extracted using an external pre-trained model.\footnote{Additionally, the visual
features are extracted by a hard-wired component, but this is a
reasonable setup at least in the context of modeling human language
acquisition: the human visual system is largely functional by the time
children start learning language in earnest.}

\paragraph{The Role of Embodied Intention}
One important source of information used by children when acquiring
language are the speaker's gaze and pointing
gestures. \citet{yu2005role} (originally presented in
\citet{yu2004multimodal}) examine the role of these clues in a
computational framework roughly similar to CELL. The most important
difference concerns the data collection process. Participants were
instructed to narrate a picture book, in English, as if speaking to a
child.  In addition to the spoken utterances and their visual context,
the authors also collected data from sensors tracking the speakers'
eye gaze, as well as head and hand position. Overall 660 utterances
were collected; average utterance size was 6 words; around 15\% of the
words in the data were object names.

The computational model features a hard-wired component which uses the
head position and gaze information to infer which part of the image
the speaker is attending to. The visual perception component extracts
image features (representing color, shape, and texture) and clusters
images into object representations. Spoken utterances are converted
into sequences of phonemes using an RNN model pre-trained on
TIMIT. Phonetically similar subsequences of phonemes are then
extracted and word-like units hypothesized based on co-occurrence with
attended-to objects. Finally, the lexicon pairing word forms with
object representations is formed by applying an alignment algorithm
based on IBM model~1 \citep{brown-etal-1993-mathematics} to the sets
of objects and sequences of words which co-occur
temporally. \todo{There are a lot of messy details also which I'm skipping.}

The performance of the system was evaluated according to four metrics:
semantic accuracy (measuring the quality of clustering of image
features into semantic groups)\footnote{Note that this is a different
  notion of semantic accuracy than that used to evaluate CELL.}, speech segmentation accuracy
(measuring whether segment boundaries correspond to word boundaries),
word-meaning association precision (the proportion of successfully
segmented words that are correctly associated with meanings) and
lexical spotting recall (the proportion of word-meaning pairs
found). The full model is compared to a baseline which ignores gaze
and head position information: the model outperforms the baseline on
the speech segmentation and especially on word-meaning association
precision, demonstrating the usefulness of inferring speaker attention
for word learning.

Most of the limitations of the CELL model also apply to
\citet{yu2005role}: the data is small-scale and the speech features
are extracted by a pre-trained model. The main innovation of the model
is the use of attention cues: the way these cues are recorded and used
by the model is not meant to match the way human learners would be
able to access and use them: rather the authors claim their
experiments as an {\it existence proof} that this information can be
usefully exploited at all.

\paragraph{The rise of deep learning}
After these initial efforts demonstrated the feasibility of learning
spoken language via visual grounding, there has been a lull in
research activity into this topic for around ten years. One important
reason for this may have been the relatively high barrier to entry:
the existing datasets were small and private, and collecting new
datasets required expertise as well as substantial time and financial
investment. Similarly, the modeling architectures relied on a lot of
custom hand-coded modules. These factors began to change with the
growing popularity of neural models after 2014: dataset collection and
sharing were crucial for the success of these approaches, and the
building blocks became standardized and available as part of
easy-to-use open-source toolkits. In the following section we will
see the effects of these developments on the research into visually
grounded models of spoken language.

\section{Datasets}
\label{sec:data}
One area on the intersection of NLP and Computer Vision which gained
popularity in the second decade of the 21\textsuperscript{st} century
was image captioning, i.e.\ the task of describing the content of
photographic images in a short written text, most commonly a single
sentence \citep{bernardi2016automatic}. The emergence of this task led to the creation of a number
of datasets designed as training data for it. The two datasets most
relevant to the current survey were Flickr8K
\citep{rashtchian-etal-2010-collecting} and COCO (also known as
MS~COCO or Microsoft~COCO) \citep{lin2014microsoft}. Both of these
datasets consisted of photographs depicting everyday objects and
situations, collected from online photo sharing services, with
accompanying textual descriptions written by crowd workers; in each
case a photo featured five independent brief descriptions. These
datasets served as a source of inspiration as well as a source of data
to enable the creation of analogous datasets for spoken language. For
the overview of the spoken datasets inspired by image captioning, see
\Cref{tab:spoken-caption-data}.

\subsection{Spoken image captions}

\begin{table}
  \centering
  \begin{small}
    \begin{longtable}{llrrrrl}
\toprule
                         Dataset & Lang. &  Images &  Captions &  Speakers &  Duration &  Speech type \\
\midrule
\endhead
\midrule
\multicolumn{7}{r}{{Continued on next page}} \\
\midrule
\endfoot

\bottomrule
\endlastfoot
           Flickr Audio Captions &    en &    8000 &     40000 &       183 &        46 &   Read aloud \\
       Synthetically Spoken COCO &    en &  123287 &    616767 &         1 &       601 &    Synthetic \\
      Synthetically Spoken STAIR &    ja &  123287 &    616767 &         1 &       793 &    Synthetic \\
                     Speech COCO &    en &  123287 &    616767 &         8 &       601 &    Synthetic \\
 Places Audio Captions (English) &    en &  400000 &    400000 &      2683 &       936 &  Spontaneous \\
   Places Audio Captions (Hindi) &    hi &  100000 &    100000 &       139 &       316 &  Spontaneous \\
                      SpokenCOCO &    en &  123287 &    605000 &      2352 &       742 &   Read aloud \\
\end{longtable}

    \end{small}
  \caption{Overview of spoken image caption datasets.}
  \label{tab:spoken-caption-data}
\end{table}

\subsubsection{Flickr Audio Captions Corpus}
The Flickr Audio Captions Corpus \citep{harwath2015deep} is based
directly on Flickr8K: it consists of the same images, while the
written captions were read aloud and recorded by crowd workers (who
were not shown the images). The recordings were quality-filtered by
running an ASR system on them and matching the output against the
original written captions and discarding the audio if more than 40\%
of the words could not be recognized. The moderate size of this corpus
(approx.~46 hours of speech) and the availability of ground-truth
written captions has made it a popular testbed for algorithmic
developments and analyses of visually grounded models of spoken language.

\subsubsection{Corpora derived from COCO}
The original COCO dataset is the source of four different spoken
caption corpora. Three of these feature synthetically generated
captions, and the third one consists of read-aloud captions.
\paragraph{Synthetically Spoken COCO}
The first of these datasets, Synthetically Spoken COCO
\citep{synthetically_2017,chrupala-etal-2017-representations} was
created by passing all the written captions of COCO through the
Google Text-to-Speech web API.\footnote{Via
  \href{https://github.com/pndurette/gTTS}{https://github.com/pndurette/gTTS}.}
The main limitation of this data is that it uses a single voice,
and lacks tempo variation, disfluencies and noise.
\paragraph{Synthetically Spoken STAIR}
This dataset \citep{william_n_havard_2018_1495070,havard2019models} is
constructed using the same methodology as Synthetically Spoken COCO
but is based on the Japanese-language version of COCO, named STAIR
\citep{yoshikawa-etal-2017-stair}. Note that the written Japanese
captions in STAIR are created independently by Japanese speakers and
not translated from English. The spoken captions are synthesized, and
exhibit the same limitations as the English synthetically spoken
captions.
\paragraph{SPEECH-COCO}
This COCO-based corpus, SPEECH-COCO
\citep{speech_coco,havard2017speech} addresses some of these
limitations. It uses Voxygen's text-to-speech system\footnote{See
  \href{https://www.voxygen.fr}{https://www.voxygen.fr}.} to
synthesize the speech for the COCO written captions, using eight
different English voices (half British, half American, half male, half
female). The tempo was manipulated: $\frac{1}{3}$ of the captions are
10\% slower than the original pace, $\frac{1}{3}$ are 10\% faster; to
30\% of the captions various disfluencies ({\it um, uh, er}) were
added.

\vskip 0.5cm These synthetic speech datasets have seen limited adoption
but have been initially useful for experiments with large-scale,
clean-speech data, and have enabled analyses which require
synthesizing stimuli (see \Cref{sec:eval}).

\paragraph{SpokenCOCO}
The final COCO-based dataset \citep{hsu2020textfree}
consists of almost all the COCO captions read aloud by 2,353 Amazon
Mechanical Turk crowd workers. To date, it has only been used in a
system to synthesize spoken captions conditioned on images
\citep{hsu2020textfree}.

\subsubsection{Places-based Corpora}
Another set of datasets is based on the images from the MIT Places 205
database \citep{zhou2014learning} which contains 2,448,873 images from
205 scene categories such as indoor (bedroom, bar, shoe shop), nature
(fishpond, rainforest, watering hole), or urban (street, tower, soccer
field).  There are two corpora based on this data: one English and one
Hindi.  The English-language dataset \citep{harwath2016unsupervised}
consists of 400,000 utterances spoken by 2,683 American English
speakers. Captions on average contain 19.3 words and have an average
duration of 9.5 seconds The Hindi-language dataset
\citep{harwath2018vision} consists of 100,000 utterances spoken by
139~Hindi speakers.  Captions contain an
average of 20.4 words and have an average duration of 11.4 seconds.

In both cases the captions are spontaneously spoken descriptions of
images from Places~205: workers are simply shown an image and asked to
describe the salient objects in several sentences.
Unlike for the Flickr8K and COCO datasets, there is only a single
caption per image. For 85,480 of the images there is a caption in both
English and Hindi.

These are the only major datasets which feature spontaneous rather
than read-aloud speech. This increases their ecological validity for
the purposes of modeling language acquisition. The flip side is that
many types of analyses become challenging due to the lack of
ground-truth transcriptions.

One point to keep in mind regarding the English Places dataset is that
results on it have been reported while it was still under development,
making direct comparisons among papers evaluating on this dataset
difficult.  

\subsection{Video datasets}
As discussed in \Cref{sec:video}, a recent development has been the
collection and use of datasets containing video clips which contain
spoken descriptions or narratives related to the activity depicted in
the video.  We briefly overview the main video-based dataset below.

\paragraph{YouCook2}
This datasets consists of approximately 2,000 cooking videos sourced
from YouTube \citep{zhou2018towards}.
\paragraph{YouTube-8M}
This is a large scale
video dataset consisting of 6.1 million YouTube videos belonging to
over 3,800 categories, including instructional videos \citep{DBLP:journals/corr/Abu-El-HaijaKLN16}.
\paragraph{Howto100m}
The dataset consists of 136 million video clips sourced from 1.22
million narrated instructional web videos \citep{miech2019howto100m}.
\paragraph{Spoken Moments in Time}
This dataset contains over 500,000 different three-second videos
depicting a broad range of different events, together with oral
descriptions \citep{monfort2021spokenmoments}.

\section{Start of the second wave}
\label{sec:first-neural}
The datasets described above eventually enabled experimentation with
the first large-scale neural models. In their pioneering work,
\citet{synnaeve2014learning} experiment with a neural
architecture trained on a dataset assembled by merging the
Pascal1K dataset \citep{rashtchian-etal-2010-collecting} which
contains one thousand images with five written captions each, together
with the LUCID speech corpus \citep{baker2010lucid}. Speech segments
corresponding to words are matched to image fragments (based on
written captions). Speech features consist of flattened Mel filterbank
coefficients; image features are extracted using a pre-trained
Regional Convolution Neural Network (RCNN) \citep{girshick2014rich}. The
model consists of a speech branch and an image branch, both of which
are mutlilayer perceptrons (MLPs) with ReLU activations. Matched and
mismatched image-word pairs are encoded into vectors by
these branches: the network is trained via the cosine-squared-cosine
loss function which ensures that matching pairs are similar while
mismatching pairs are orthogonal in the vector space:
\begin{equation}
  \label{eq:coscos2}
  \ell_{\textsc{coscos}^2} = \begin{cases}
    \frac{1}{2}(1-\cos(I, S)) & \text{ if matched }\\
    \cos^2(I, S)              & \text{ if not matched }
    \end{cases}
\end{equation}
where $I$ and $S$ are the vectors corresponding to the image fragment
and spoken word, respectively, and $\cos(\cdot, \cdot)$ is the cosine similarity
function. The model is evaluated on retrieving words (types or tokens)
given an image and vice versa, and compared to a random baseline, which
it consistently outperforms. This work is the first of the second
wave, but quite preliminary in its use of a small and
artificially assembled dataset.

\paragraph{Embedding alignment model}
Independently\footnote{The work of \citet{harwath2015deep} seems to be
  independent, as \citet{synnaeve2014learning} is not cited.},
\citet{harwath2015deep} introduced the Flickr Audio Captions dataset
and reported on experiments with a model to learn {\it multimodal
  embeddings} of this data using a somewhat similar neural
architecture. The images were processed using an RCNN pre-trained in a
supervised fashion on ImageNet, extracting 4096-dimensional
feature vectors for each of 19 regions in an image, plus one for the
whole image. The spoken utterances are preprocessed by segmenting the
audio at word boundaries via force-alignment to ground-truth
transcriptions. The spectrograms corresponding to words were then
passed through a supervised CNN model pre-trained on the Wall Street
Journal SI-284 split \citep{paul1992design} and using the
1024-dimensional activation vectors from the final fully connected
layer as audio features.

The core of the architecture is the {\it
  embedding alignment model}, based on
\citet{karpathy2014deep}, whose job it is to embed the
features extracted from audio segments and region images into a common
representation space. The visual features are projected to this space via an
affine transform, while the audio features are projected via an affine
transform followed by a ReLU. The objective function used to train
these transforms is the triplet-like loss from \citet{karpathy2015deep}: a
version of this loss is used in most subsequent work on visually
grounded models of spoken language. Its general form is defined as:

\begin{equation}
\ell = \sum_{ui}\left[\sum_{u'} \max(0, S_{u'i} - S_{ui} +
  \alpha) + \sum_{i'} \max(0, S_{ui'} - S_{ui} + \alpha) \right]
\label{eq:triplet}
\end{equation}
where $\alpha$ is a margin, $S_{ui}$ is a similarity score between
a matching caption-image pair, and $S_{u'i}$ and $S_{ui'}$ denote
similarity scores between mismatched pairs, i.e.\ negative examples
(typically sampled from the current batch).  In
\citet{harwath2015deep} the margin is set to $\alpha=1$ and the caption-image
similarity is measured by:
\begin{equation}
  \label{eq:summax}
  S_{ui} = \sum_{t \in \mathrm{regions}(i)} \max_{i \in \mathrm{words}(u)}(0, e_i^Te_t),
\end{equation}
where $e_i^Te_t$ stands for the inner product between the embedding of
word $i$ and the embedding of region $t$.
In order to evaluate the architecture, \citet{harwath2015deep} adopt
the approach from work on written captions and rank images with
respect to a caption (or captions with respect to an image) and report
recall@10, i.e.\ the proportion of true matches that are found
in the top ten items in the ranking. See
\Cref{sec:intrinsic} for details of the evaluation metrics and
for an overview of the results on the different datasets.

\paragraph{Predicting image features from phoneme strings}
\citet{gelderloos-chrupala-2016-phonemes} experimented with COCO
images and written captions converted into sequences of
phonemes:\footnote{See \Cref{sec:phonology} for a brief discussion of
  phonemes and phonemic transcription.} as such their work is not
strictly speaking about modeling spoken language in its audio
modality\footnote{Note that while a few other works mentioned here
  resort to transcribing spoken utterances using a pre-trained ASR
  model, \citet{gelderloos-chrupala-2016-phonemes} do not work with
  the speech signal at all but rather convert the textual captions to
  canonical phonemic transcriptions directly.}; it is, however, worth
mentioning here because it introduces many of the approaches to
analysis of learned representations which became a major preoccupation
in later work with visually grounded models of spoken language.

The phonetic transcriptions are automatically generated from textual
captions using the grapheme-to-phoneme mode of
eSpeak\footnote{Available at
  \href{http://espeak.sourceforge.net}{http://espeak.sourceforge.net}};
spaces are removed to simulate connected speech. The visual part of
the architecture consists of a convolutional neural network
\citep{simonyan2015deep} pre-trained on the ImageNet database
\citep{deng2009imagenet}: visual features for an image are obtained by
extracting the activations of the final fully-connected layer in this
network. The phonetic input was processed via a stack of three Gated
Recurrent Unit (GRU) layers \citep{chung2014empirical} (with residual
connections between layers). The activation vector of the top layer at
the last timestep is then mapped to match the image feature vector and
learning proceeds by minimizing the mean squared error (MSE) of the
predicted image vector. The focus of this work is on understanding
how the network learns to construct a series of internal
representations which enable it to map sequences of phonemes to visual
features: it shows that lower layers in the recurrent stack encode
comparatively more information related to form, whereas higher layers
encode meaning more strongly. This conclusion is reached based on
analyses such as predicting word boundaries from recurrent layer
activations (which was easiest for layer~1) as well as correlating
word similarities obtained from the network activations to
phoneme-string edit distances (the relation is strongest for layer~1)
and to human similarity judgments (the correlation is highest for
layer~3). These analyses pre-figured those used in recent work on
analyzing neural models of (spoken) language: see also \Cref{sec:extrinsic}.

\section{Encoder architectures}
\label{sec:arch}
The basic template of models of visually grounded spoken language
consists of a pre-trained image classification network used to extract
image features, together with an encoder for the audio signal, trained end-to-end,
combined with the module which matches these two modalities such as
the embedding alignment model described in
\Cref{sec:first-neural}. Most of the architectural variability
has been on the side of the audio encoder.
\Cref{fig:implement} shows the most prominent software implementations
of these architectures, with a link to the repository and a brief
comment on the main features.

\todo{Say something about visual features also?}

\subsection{Convolutional audio encoders}
\label{sec:conv}
Many works have followed \citet{harwath2016unsupervised} in using a
convolutional architecture with a spectrogram as input to encode the
audio
signal \citep[e.g.][]{harwath-glass-2017-learning,harwath2018jointly,harwath2018vision,boggust2019grounding}.
%\todo{Any more papers using this?}
This approach treats the input as a two-dimensional
grayscale image, and adapts convolutional architectures from computer
vision to the work with spectrograms: the encoder consists of a number
of two-dimensional convolutional layers interspersed with maxpooling
layers. In order to capture the fact that the vertical dimension in a
spectrogram which corresponds to frequency should not be translation
invariant, the height of the first convolutional layer spans the
entire height of the spectrogram (40 pixels, each pixel corresponding
to one of 40 bandpass filters), thus collapsing the frequency
dimension such that subsequent layers are convolutional only over the
time dimension. The final maxpooling operation spans the entire
duration of the utterance and aggregates it into a single vector,
which is then L2-normalized. In practice, the spectrogram is also
truncated or padded such that all utterances are of the same
size, corresponding to 10s or 20s. As an example,
\Cref{tab:encoder-conv} shows the full specification of the audio
encoder as described in \citet{harwath-glass-2017-learning}.
\begin{table}\centering
  \begin{tabular}{rlrrrrl}\toprule
    
       & Layer       & Channels & Width & Height & Stride & Activation \\\midrule
     1 & Convolution & 128      &  1    & 40     & 1      & ReLU \\
     2 & Convolution & 256      & 11    &  1     & 1      & ReLU \\
     3 & Maxpool     &          &  3    &  1     & 2      &      \\
     4 & Convolution & 512      & 17    &  1     & 1      & ReLU \\
     5 & Maxpool     &          &  3    &  1     & 2      &      \\
     6 & Convolution & 512      & 17    &  1     & 1      & ReLU\\
     7 & Maxpool     &          &  3    &  1     & 2      &     \\
     8 & Convolution & 1024     & 17    &  1     & 1      & ReLU \\
     9 & Meanpool    &          & $\infty$&  1     & 1      &      \\
    10 & L2 norm     &          &       &        &        &      \\\bottomrule
  \end{tabular}
  \caption{The architecture of the convolutional speech encoder in
    \citet{harwath-glass-2017-learning}.}
  \label{tab:encoder-conv}
\end{table}

\paragraph{Residual Network}
Another technique borrowed from computer vision is the use of residual
blocks which belong to the same general family of convolutional layers
but with some important innovations, especially the introduction of
residual connections.
%This operation decomposes the output
%$y$ of a residual block as $y = F(x)+x$ with $F$ being the residual
%mapping to be learned; this tends to facilitate the training of deeper
%network architectures.  
Such a {\it ResNet} encoder is described in 
\citet{hsu2019transfer} with the make-up of the residual blocks
borrowed from \citet{he2016deep}, and further adapted to induce
discrete units via vector quantization in \citet{harwath2020learning}.

\subsection{Recurrent audio encoders}
\label{sec:recur}
The main alternative to a convolutional speech encoder has been to use
some kind of a recurrent architecture applied to the input converted
to Mel-Frequency Cepstral Coefficient (MFCC) representation (with
delta and double-delta features, which capture the first and second
derivative of the converted signal).
\citet{chrupala-etal-2017-representations} introduce an encoder which
consists of a convolutional layer followed by a stack of recurrent
highway network (RHN) layers \citep{zilly2017recurrent}, followed by an
attention-like pooling operator.\footnote{RHNs are a type of recurrent
  network which feature multiple so-called micro-steps between actual
  times-steps of the input. Subsequent work
  \citep[e.g.][]{chrupala-2019-symbolic,Merkx2019} has tended to
  replace RHNs with simpler layers such as (bidirectional) GRUs which
  have widely available optimized low-level CUDA support, making them
  much more efficient.}  The initial convolutional layer used here is
a one-dimensional convolution along the time dimension, with a stride
$>1$ resulting in a temporal subsampling of the signal. The pooling
operator in this family of architectures is a weighted sum over the
full extent of the time dimension, where the weights are learned (also
called self-attention). The
original recurrent encoder of
\citet{chrupala-etal-2017-representations} features scalar weights
computed by applying a Multi-Layer Perceptron to the activations of
the topmost recurrent layer, resulting in a single weight per
timestep. More recent work
\citep[e.g.][]{Merkx2019,chrupala-etal-2020-analyzing} has used
vectorial weights, resulting in a single
weight per time-step per dimension, defined as:
\begin{equation}
  \label{eq:attention}
  \mathrm{Attn}(\mathbf{h}) = \sum_t \boldsymbol\alpha_t \odot \mathbf{h}_t
\end{equation}
with the weight vectors computed via:
\begin{equation}
  \label{eq:attention-w}
  \boldsymbol\alpha_t = \mathrm{softmax}_t \left(\mathrm{MLP}(\mathbf{h}_t)\right)
\end{equation}
As an example, \Cref{tab:encoder-recurrent} shows the full
specification of the recurrent encoder as described in \citet{chrupala-etal-2020-analyzing}.
\begin{table}\centering
  \begin{tabular}{rll}\toprule
    & Layer                & Parameters   \\\midrule
    1  & 1D Convolution    & Channels=64,  Kernel=6,  Stride=2 \\
    2  & Bidirectional GRU &  Hidden=$2\times 1024$, Layers=4 \\
    3  & Attention pooling & Hidden=128, Vectorial        \\\bottomrule
  \end{tabular}
  \caption{The architecture of the recurrent speech encoder in
    \citet{chrupala-etal-2020-analyzing}.}
  \label{tab:encoder-recurrent}
\end{table}

\subsection{Temporal and spatial localization}
\label{sec:timespace}
All the architectures described in \Cref{sec:conv,sec:recur} rely on
encoding both the utterance and the image into a single vector, thus
aggregating over the time and space dimensions and making the models
incapable of associating regions of the images to spans of the audio
signal. This limitation is addressed in \citet{harwath2018jointly} by
discarding the final pooling and fully connected layers of the visual
encoder. The feature maps in the final convolutional layer of the
resulting network can be directly related to the input
image. Similarly, the audio encoder is modified such that it does no
subsampling or pooling over the whole extent of the caption, and thus
keeps temporal localization. The relations between regions in the
image feature maps and the frames of the audio feature map are encoded
in the so-called \textit{matchmap} which is a three-dimensional tensor
storing affinities between each point in the two-dimensional visual
feature map and each frame in the one dimensional feature map of the
audio signal. The values in this matchmap are then aggregated into an
overall image-utterance similarity score needed by the objective in
\Cref{eq:triplet}, via a scoring function. The one which tends to
perform best is named MISA (for maximum over image, sum over audio)
and it matches each frame in the audio feature map to the best image
patch, and then averages over the whole utterance:
\begin{equation}
  \label{eq:misa}
  \mathrm{MISA}(M) = \mean_t \left(\max_{r,c} M_{r,c,t} \right),
\end{equation}
where $M$ is the matchmap, $N_t$ is the number of audio frames,
and $r,c$ range over the spatial coordinates of the visual feature
map.

\subsection{End-to-end visual encoder}
\label{sec:end-to-end}
Unlike most previous and subsequent work, \citet{harwath2018jointly}
 experiment not only with a pre-trained visual encoder, but also
evaluate the performance of one trained from scratch only on the
images in the Places dataset. As expected due to limited training
data, the performance in this condition is substantially lower (see
\Cref{tab:caption-retrieval-places,tab:image-retrieval-places}), but
the experiment shows that it is feasible to train a visually grounded
model of spoken language entirely end-to-end.

\begin{figure}
  \centering
  \fbox{\parbox{\textwidth}{
\begin{description}
\item[DAVEnet]~ \\
  \citet{harwath2018jointly}\\
  \url{https://github.com/dharwath/DAVEnet-pytorch} \\
  Audio is converted to a spectrogram and processed via a CNN.
\item[ResDAVEnet-VQ]~ \\
  \citet{harwath2020learning}\\
   \url{https://github.com/wnhsu/ResDAVEnet-VQ} \\
   Audio is converted to a spectrogram and processed via a ResNet, 
   with vector-quantization layers inducing discrete representations.
 \item[Platalea]~\\
   \citet{chrupala-etal-2020-analyzing} \\
   \url{https://github.com/spokenlanguage/platalea} \\
   Audio is converted to MFCC features and processed via a stack of
   bidirectional GRUs, followed by attention pooling.
\item[Speech2image]~ \\
  \citet{Merkx2019}\\
  \url{https://github.com/DannyMerkx/speech2image} \\
  Audio is converted to MFCC as well as a number of other audio feature types
  and processed via a stack of bidirectional GRUs, followed by attention pooling.
\end{description}
}}

\caption{Overview of implementations of models of visually grounded spoken language learning.}
\label{fig:implement}
\end{figure}

\section{Grounding in video}
\label{sec:video}
The majority of extant work on visual grounding of spoken language has
used static images, due both to the wider availability of
curated captioned image datasets, as well as the inherent complexity
of modeling two weakly synchronized modalities in the temporal
domain. However, for humans visual perception is inherently extended
in time, and visual grounding of language based on still images is
arguably suboptimal also from an engineering point of view.  Many
aspects of language are related to processes or actions evolving over
time rather than static states and may be more naturally grounded in
video. This intuition has led to increasing interest in collecting
video datasets coupled with spoken narratives and extending the
modeling approaches to the video modality. This fast-developing area
will likely see important advances in the immediate future: we thus do not
attempt a detailed, definitive account of current modeling efforts
here, but rather choose to provide a concise high-level overview.

\paragraph{Single video frames} \citet{boggust2019grounding} tackle
this challenge by focusing on the constrained domain of instructional
cooking videos and explore the various degrees of supervision applied
to sampling audio-visual fragments. The architecture itself does not
attempt to model the temporal nature of video and simply applies
DAVENet to stills extracted from the video stream together with
samples of audio. Even with no supervision (i.e.\ uniform sampling of
still-audio pairs) the model does learn some cross-modal
correlations. The loose synchrony between the two modalities, such
that objects may be mentioned in the audio at a different point in
time than they occur in the video, remains the main challenge for this
approach.

\paragraph{Video encoder} \citet{rouditchenko2020avlnet} present an
architecture (AVLNet) which does model the time dimension in the video
stream: the network consists of an audio encoder (ResNet-based), a
video encoder which combines 3D and 2D modeling (also ResNet-based),
as well as an optional text encoder. This architecture is trained with
a contrastive loss on randomly sampled audio-video
fragments from the Howto100m dataset \citep{miech2019howto100m}
consisting of 136 million video clips sourced from 1.22 million
narrated instructional web videos.  The model is evaluated on the
video clip and language retrieval tasks on smaller video datasets
annotated with clip boundaries and text summaries, and is shown to
outperform previously proposed models of \citet{arandjelovic2018objects} and \citet{boggust2019grounding}.
The model also transfers to the image-audio retrieval setting.
Qualitative analysis suggests that
the model aligns semantically related audio and visual features to
particular dimensions of the embedding space.

\paragraph{Beyond instructional videos} The instructional videos used
in the above-mentioned works are very specific and limited in the type
of visual scenes and language they contain. These limitations are
lessened in the Spoken Moments in Time dataset which contain over
500,000 different three-second videos depicting a broad range of
different events, together with oral descriptions
\citep{monfort2021spokenmoments}. The authors show that a model
trained on this dataset does indeed generalize better than those
trained on other video-caption data, due to its large coverage,
diversity and scale. To enable this cross-dataset comparison, an
architecture containing a pre-trained ASR module was used, but in
addition a model directly aligning spoken captions with video was
likewise shown to be effective.

\section{Variants and applications}
\label{sec:variations}

The basic architectures described above have given rise to many
variations tackling different additional aspects of visual grounding
of spoken language. These include application to keyword spotting and
speech-based retrieval,
multilingual models, the use of auxiliary supervision, fine-grained
localization and language unit discovery.

\paragraph{Keyword spotting}
One line of work, inititated by \citet{kamper2017visually}, uses the
visual modality as a bridge to learn keyword spotting from
untranscribed but visually grounded speech. The architecture consists
of a (convolutional) audio encoder which is trained via the cross
entropy loss to mimic the output of the softmax over labels of a
pretrained image classification model, for the image paired with the
spoken utterance. This enables the model, after training, to map
spoken utterances to a bag-of-words representations, where the words
are the labels used by the image classification system. This model can
then act as a keyword-based retrieval method, where the goal is to
find all the utterances in a candidate pool containing the keywords in
a given query. The authors show that the mistakes made by this system
are often based on meaning and not the sound of the words, for
example mixing up the keywords {\it boys} and {\it children}. This
makes the method effective as a semantic keyword spotter. This work
has led to a number of follow-ups and extensions 
\citep[e.g.][]{kamper2018visually,kamper2019semantic,
  kamper2019semantic-ieee,olaleye2020localisation}.

\paragraph{Multilinguality}
The large majority of work on visually grounded models of spoken
language have focused on a single language, and specifically on
English. The most salient exception is \citet{harwath2018vision} who
collect Hindi spoken captions for a subset of the Places images, and
use this data and explore a multilingual architecture based on
\citet{harwath-glass-2017-learning}, which projects images, English
speech and Hindi speech into a joint semantic space via three separate
encoders. They experiment with loss functions which combine ranking
objectives in various configurations: for example rank English
captions against an image, rank English captions against a Hindi
caption, rank Hindi captions against an image, etc. They find that a
multilingual model trained on both languages outperforms monolingual
models, and also show the feasibility of semantic cross-lingual
speech-to-speech retrieval using a multilingual model.

\citet{havard2019models} create a dataset of synthetic Japanese
captions, and show that visually grounded models of spoken language
based on the recurrent encoder architecture and trained on English and
Japanese captions learn to focus the self-attention weights
on nouns more than to any other category of
word. Additionally, the Japanese model also focuses on particles such
as \textit{ga} which indicate grammatical function, mimicking Japanese
toddlers in this regard.

\citet{ohishi2020trilingual}
collect a Japanese version of the Places audio captions in order to
explore tri-lingual models and further \citet{ohishi2020pair} propose
pair-expansion techniques for settings when captions are not aligned
across languages. This dataset is not currently publicly available but
the authors plan to release it in future.\footnote{David Harwath,
  personal communication.}

\citet{kamper2018visually} use the visual modality as a pivot to
enable a cross-lingual task. In their case it is cross-lingual
keyword-spotting: given a text keyword in one language, the task is to
retrieve spoken utterances containing that keyword in another
language. They adapt the approach of \citet{kamper2017visually} to the
setting of retrieving English spoken utterances using German text
keywords.

\paragraph{Auxiliary textual supervision}
Several works have proposed including additional supervision signal
into the basic visually-grounded speech scenario using textual data in
some form. This idea departs somewhat from the premise of most other
work discussed so far in that it relies on textual supervision to some
extent. In some settings however, the availability of moderate amounts
of text is a reasonable assumption.  \citet{chrupala-2019-symbolic}
demonstrates the use of speech transcriptions in a multi-task setup as
a way of injecting an inductive bias to nudge the model towards
learning more symbol-like representations.  \citet{Pasad2019} use a
very similar approach while specifically focusing on low-resource
settings and testing the effect of varying the amount of textual
supervision. \citet{higy2020textual} investigate two forms of textual
supervision in low-resource settings: transcriptions, and text
translations. They also compare the multi-task learning approaches to
simple pipeline architectures where text transcriptions are used to
train an ASR module, and find that in most cases the pipeline is hard
to improve on. \citet{ilharco-etal-2019-large} exploit textual
supervision in a more indirect manner, via a text-to-speech
(TTS) system trained on transcribed speech. The use it to automatically generate
large amounts of synthetic training material for a visually grounded
speech model: they obtain substantial performance improvements using
this approach.

\paragraph{Fine-grained spatial localization}
\citet{PontTuset_eccv2020} ask annotators to orally describe an
image while hovering their mouse over a relevant region of the
image. This results in each audio segment being explicitly grounded in
a specific portion of the image. The resulting dataset, Localized Narratives,
consists of 849,000 images sourced from COCO, Flickr30k, and
ADE20K \citep{zhou2019semantic} and Open Images
\citep{kuznetsova2020open}. \citet{PontTuset_eccv2020} demonstrate the
utility of this information for controlled image
captioning in the written modality. However, it is clear that such alignments between
audio and image regions would also be useful as an auxiliary
supervision signal for visually grounded models of spoken language, as well
as for analysis and evaluation of these architectures. A related
dataset, with a similar potential, is Room-across-Room which has
spoken navigation instructions that are temporally aligned with the
guide's 3D camera pose \citep{ku-etal-2020-room}.

\paragraph{Unit discovery}
An emerging trend in neural architectures, especially as applied to
the speech signal, is the use of mechanisms to enable them to induce
discrete, symbol-like internal representations, motivated by concepts
such as phonemes and morphemes. Recent editions of the ZeroSpeech
challenge \citep{dunbar2019zero} on unit discovery have featured many
such approaches. In the visually-grounded setting,
\citep{harwath2020learning} adopt the vector-quantization (VQ)
approach proposed by \citet{van2017neural}, inserting VQ layers at
various points in the speech encoder. They analyze the nature of the
learned discrete units. They see evidence that units at lower
levels correspond roughly to phonemes while those at higher levels are
more word-like. \citet{DBLP:journals/corr/abs-2105-05582} carry out detailed
analyses of a visually grounded model with VQ similar to
\citep{harwath2020learning} as well as a self-supervised speech-only
model of \citet{van2020vector}, testing the effect of codebooks size
(i.e.\ the number of discrete units), as measured via several
metrics. They show that the different evaluation metrics can give
inconsistent results, and that while in general VQ-based discrete
representations do correlate with units posited in linguistics, this
correlation is moderate in strength at best.

\paragraph{Speech-based image retrieval}
Retrieval of images based on spoken captions often features as one of
the motivations for visually grounded models of spoken language; image
and/or caption retrieval has also served as the basis of the intrinsic
evaluation metrics for these models (see \Cref{sec:intrinsic}). In
contrast, \citet{sanabria21_interspeech} focus on speech-based image
retrieval as \textit{the} application of interest and carry out an extensive
study of how this task is affected by encoder architectures and
training and pre-training methodology. Their best configuration
outperforms an ASR-based pipeline approach in cases when speech is
spontaneous, accented, or otherwise challenging.

\citet{peng2021fastslow} also focus on speech-based retrieval and
propose an architecture which combines two modality-specific encoders
for speech and images with dual-modality featuring cross-modality
attention.\footnote{Transformer is a type of network which relies on
  several layers of multiple self-attention heads to encode input
  efficiently; this component was first introduced for NLP tasks
  \citep{vaswani2017attention} and has since been applied to many
  other domains: \citet{peng2021fastslow} is the first work to adopt it
  for modeling
  visual grounding of speech.} The single-modality encoders enable
fast retrieval of matching candidates, as the encoding of each input
can be computed independently and cached. The dual-modality encoder
with cross attention needs to be applied to each pair of inputs and is
thus slow, but more accurate: it is thus applied to a small set of
candidates retrieved via the single-modality encoders, resulting in an
overall fast and accurate retrieval. The system is evaluated on
multiple image caption datasets (Flickr Audio Captions, Places and
SpokenCOCO); the representations learned are benchmarked against
datasets provided by the 2021 Zero-Resource Speech Challenge, Visually Grounded Track
\citep{alishahi2021zr2021vg}.

\paragraph{Image generation}
\citet{wang2021learning} investigate the learning of fine-grained
visual distinctions from two datasets: one of bird specimens
\citet{wah2011caltech} and one of flowers
\citet{nilsback2008automated}.  In both cases spoken captions are
generated using a TTS system. Fine-grained visual representations are
evaluated cross-modal retrieval as well as speech-to-image generation,
implemented using an architecture based on
\citet{zhang2018stackgan++}.

\paragraph{Unwritten languages}
One practical motivation for developing visually grounded models of
spoken language has been to enable speech-based applications for
language without a widely-used standardized writing system. For such
language the use of standard ASR and Text-to-Speech (TTS) approaches
is difficult due to lack of transcribed
speech. \citet{scharenborg2020speech} focus on this setting and
describe systems to carry out translation of speech, speech-to-image
retrieval, and image-to-speech generation. The first system is trained
on data from a true unwritten language (Mboshi), while the latter two
rely on visual grounding via the Flickr Audio Caption dataset,
used to simulate the unwritten language scenario.

\section{Evaluation metrics and analysis techniques}
\label{sec:eval}
This section focuses on evaluation and analysis techniques for the
second-wave systems. The evaluation of early models has some
parallels, but the metrics are often very model-specific and are thus
discussed in \Cref{sec:early}.

The most widely used approach to measure the performance of visually
grounded models of spoken language has been to use a set of metrics
closely related to these model's objective function, i.e.\ the triplet
loss from \Cref{eq:triplet}, but more clearly interpretable. We will
refer to this set of metrics as \textit{intrinsic}.

In addition, many works have proposed a variety of ways to analyze and
evaluate the spoken utterance embeddings extrinsically, with reference
to some external gold standard representations of phonological form,
the lexicon or semantics. We will refer to these types of evaluations
as \textit{extrinsic}.
\subsection{Intrinsic evaluation}
\label{sec:intrinsic}
%NOTE: The retrieval scores numbers in
%\citet{harwath2016unsupervised} and
%\citet{harwath-glass-2017-learning} different, most likely due to the
%data used in the former being a subset of the latter:
%\begin{itemize}
%\item \citet{harwath2016unsupervised} 114,000 train, 2,400 dev, 2,400
%  test
%\item \citet{harwath-glass-2017-learning} 214,585 train, (no dev),
%  1,000 test
%\end{itemize}

The main metrics to evaluate intrinsic performance are based on
setting up a retrieval task on the development and/or test set
image-caption pairs.  The learned embeddings are used to rank images with
respect to a given caption, or rank captions with respect to a given
image, and we measure the quality of the ranking by checking whether
the true matches are top-ranked. For clarity, in the following we will
refer to retrieving images associated with a give caption; the metrics
work analogously for retrieving captions for an image.

Specifically, \textit{recall@N} refers to looking at the top $N$
images in the ranking, and calculating the proportion of the truly
matching images captured within this set, averaged over the captions.
\textit{Median rank} refers to the position in the ranking of the
truly matching image, taking the median of this number over the
captions.  A true match is determined by whether the caption and image
are associated in the dataset: it should be noted that using this
definition of true match will tend to underestimate the performance of
the embeddings, as it is possible and not uncommon for multiple other
images in the data to also be semantically closely associated with the
given caption, even though they are not paired in the dataset. This
can arise because there may be multiple images depicting quite similar
situations.

According to these metrics, there has been substantial progress on the
two most widely used datasets, Flickr Audio Captions and
Places. \Cref{tab:image-retrieval-flickr8k} shows the results for
image retrieval on Flickr as reported in a number of works since
2015 (caption retrieval numbers are not available for most of these
systems). \Cref{tab:caption-retrieval-places,tab:image-retrieval-places}
show the results for both caption and image retrieval on Places: note
that here most of the numbers come from the experiments reimplemented
on the complete data as reported by \citet{harwath2018jointly}, as the
numbers reported in previous papers used different subset and/or split
of the dataset.

\begin{table}
  \centering
  \begin{tabular}{lrrrr}
    \toprule
    System                  & Recall@1 & Recall@5 & Recall@10 & Median rank \\\midrule
    \citet{harwath2015deep} & -        & -        & 0.179     & -     \\
    \citet{chrupala-etal-2017-representations}
                            & 0.055    & 0.163    & 0.253     & 48    \\
    \citet{Merkx2019}       & 0.084    & 0.257    & 0.376     & 21    \\
    \citet{scholten2020learning}
                            & 0.107    & 0.292    & 0.402     & 18    \\
    \bottomrule
  \end{tabular}
  \caption{Overview of image retrieval performance on Flick Audio Captions test data.}
  \label{tab:image-retrieval-flickr8k}
\end{table}

\begin{table}
  \centering
  \begin{tabular}{lrrr}
    \toprule
    System                   & Recall@1 & Recall@5 & Recall@10  \\\midrule
%    Harwath 2016 (max,20)    & 0.068    &  0.223   & 0.309     \\
%    Harwath 2017 (2016)      & 0.090    &  0.261   & 0.372      \\
%    Harwath 2017             & 0.112    &  0.312   & 0.431       \\
    \citet{harwath2016unsupervised}     & 0.148     & 0.403     & 0.548     \\
    \citet{harwath-glass-2017-learning} & 0.161     & 0.404     & 0.564    \\
    \citet{harwath2018jointly}   MISA         & 0.200     & 0.469     & 0.604     \\  
    \citet{harwath2018jointly}   MISA end-to-end       & 0.079     &
                                                                     0.225     & 0.314    \\
    \citet{khorrami_2021} CNN1 & -  & - & 0.522 \\
    \bottomrule
    \end{tabular}
    \caption{Overview of caption retrieval performance on Places test
      data, with models trained on full Places training data. First
      four rows as reported in \citet{harwath2018jointly}.}
  \label{tab:caption-retrieval-places}
\end{table}

\begin{table}
  \centering
  \begin{tabular}{lrrr}
    \toprule
    System        & Recall@1 & Recall@5 & Recall@10  \\\midrule
%    Harwath 2016 (max,20)   & 0.061    & 0.192    & 0.291     \\
%    Harwath 2017 (2016)     & 0.098    & 0.266    & 0.352      \\
%    Harwath 2017            & 0.120    & 0.307    & 0.438      \\
    \citet{harwath2016unsupervised}      & 0.121     & 0.335    & 0.463      \\
    \citet{harwath-glass-2017-learning}  & 0.130     & 0.378    & 0.542      \\
    \citet{harwath2018jointly} MISA      & 0.127     & 0.375    & 0.528      \\
    \citet{harwath2018jointly} MISA end-to-end  & 0.057     & 0.191
                                        & 0.291       \\
    \citet{khorrami_2021} CNN1          & - & - & 0.525 \\
    \bottomrule
  \end{tabular}
  \caption{Overview of image retrieval performance on Places test
    data, with models trained on full Places training data. First four
    rows as reported in \citet{harwath2018jointly}.}
  \label{tab:image-retrieval-places}
\end{table}

\subsection{Extrinsic evaluation}
\label{sec:extrinsic}
Neural network architectures such as those adopted for the recent work
on visually grounded models of spoken language use many levels of
hidden distributed representations. This often presents an obstacle
when we want to understand what the models have learned and to control
some aspect of their behavior. The solution has been to devise
techniques for probing these representations and analyzing their relation
to established units and representations from linguistics.
\subsubsection{Phonological form}
\label{sec:phonology}
For the purpose of linguistic analysis the form of spoken utterances
is often abstracted into a sequence of basic units known as
phonemes. Phonemes are the smallest sound units specific to a language
such that exchanging one phoneme for another can alter
the meaning of the utterance, as in the following pair of English
words: \textit{pad-pat}. \Cref{tab:ipa} shows an example utterance
rendered in standard English spelling and a sequence of
phonemic symbols.

Phonological form abstracts away much of the information present in
the concrete waveform of an utterance, including speaker identity and
demographic attributes, tone of voice including emotion, tempo, or
environmental noise. We would expect that a visually grounded model
trained on spoken image captions would also learn to abstract away
these aspects of the audio signal, given that representing them is not
useful to score well on the model's objective function.  Additionally,
if linguistic theory is correct in positing that utterances are built
up from phonemes, we may wonder whether models are able to discover
this fact. For these reasons several works have proposed ways of
testing the learned representations of visually grounded models of
spoken language against ground-truth phonological forms; similar
analyses have also been carried out for automatic speech recognition (ASR)
models \citep{belinkov2017analyzing,krug2018neuron,Belinkov_2019}.
\begin{table}
  \centering
  \includegraphics{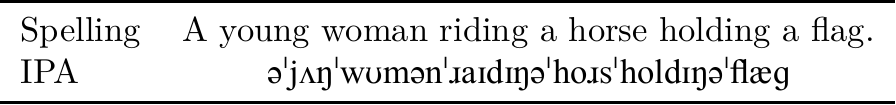}
  \caption{Phonological form (US-English) in the International
    Phonetic Alphabet for an example utterance. The symbol \textipa{"}
  indicates that the following syllable is stressed.}
  \label{tab:ipa}
\end{table}

Analyzing and evaluating neural representations is a rapidly changing
field and as such there are a variety of methodological proposals and
results and there is no consensus so far on what the best practice
should be. Below we outline the most prominent methods which have been
applied to evaluating phonological forms in visually grounded models
of spoken language. Note that most of these methods are quite generic
and can be applied to units or representations other than phonological
form, but here we focus on their use to probe for phonological information. 

\paragraph{Phoneme boundary detection}
\citet{harwath2019towards} extract activations of the {\tt conv2}
layer of the model of \citet{harwath2018jointly}, correlate peaks in
activation (after applying L2-norm across channels) with ground-truth
phoneme boundaries in TIMIT, and find that the resulting phone
boundary detection accuracy compares favorably with existing
unsupervised methods. This suggests that utterances are implicitly
segmented into phonemes within this architecture. These findings were
partially corroborated by \citet{khorrami_2021}, with the proviso of
rather lower scores and the fact that implicit phoneme segmentation is
also present to a large extent in activations from untrained models,
and thus is not fully due to learning, but simply to network dynamics.

\paragraph{Diagnostic classifier (DC)}
A common approach is to train a separate diagnostic model (or probe) which
takes the representation learned by the target visually grounded model
as input features and ground-truth phoneme labels as labels. This
model's accuracy on held-out data can be seen as a measure of the
strength of encoding of phonology in the target representations.
\citet{alishahi-etal-2017-encoding} apply this idea to the model of
\citet{chrupala-etal-2017-representations} trained on the
Synthetically Spoken COCO dataset. Overall, they find that phoneme
representations are most salient in lower layers of the architecture,
although they do persist even in the top layers.  
\citet{chrupala-etal-2020-analyzing} recommend to compare
the accuracy of the diagnostic classifier on the target model to the
accuracy on the untrained version of the target model: it is often the
case that even the representations extracted from an untrained,
randomly initialized model contain enough information to enable the
diagnostic classifier to perform substantially above the majority
baseline. 

\paragraph{ABX}
The ABX discriminability metric was introduced by
\citet{schatz2016abx} and used by
\citet{alishahi-etal-2017-encoding} to evaluate phoneme
representations in the model of
\citet{chrupala-etal-2017-representations}. It is based on triples
of stimuli ($A, B, X$) where $A$ and $X$ belong to the same category
and $B$ and $X$ belong to different categories. The ABX error is a
function of $d(A, X)$ and $d(B, X)$ where $d(\cdot, \cdot)$ is a
distance metric for the representation being evaluated:
\begin{equation}
  \label{eq:abx}
  \mathrm{abx}(A, B, X) =
  \begin{cases}
    1           & \text{ if } d(A, X) > d(B, X) \\
    \frac{1}{2} & \text{ if } d(A, X) = d(B, X) \\
    0           & \text{ otherwise }
  \end{cases}
\end{equation}
In the case of phoneme discrimination, the categories are determined
by ground-truth phoneme information: consider the case where
$A=\textit{tu}$, $B= \textit{du}$ and $X= \textit{ti}$.
The task is to determine whether \textit{ti} should be grouped with
target \textit{tu} or with \textit{du}. The correct answer is
\textit{tu} since it only differs from \textit{ti} by one phoneme and
they form a minimal pair, while \textit{ti} and
\textit{du} are not a minimal pair. Note that the target and
distractor are typically matched on some attribute, in this case the
context vowel. When applied to representations such as sequences of
activation vectors, matching between stimuli is determined by
computing pairwise distances. This can be done by first mean- or max-pooling the
activations, and then calculating the cosine (or other) distance
between the resulting vectors, or by directly computing the distance
between two sequences of vectors using the dynamic time warping (DTW)
algorithm.  
One feature of the ABX metric is that it is based on a
set of tightly controlled stimuli. \citet{alishahi-etal-2017-encoding}
used synthetic audio; in other work such stimuli have been extracted
from utterances using aligned phonemic transcriptions
\citep{dunbar2019zero,DBLP:journals/corr/abs-2105-05582}.

\paragraph{RSA}
Representational Similarity Analysis (RSA) originates in neuroscience
\citep{kriegeskorte2008representational} where similarities or
distances between pairs of stimuli are computed in two representation
spaces: in our case these would be the activation space and the space
of phonological forms. The correlation between these pairwise distance
measurements shows the degree of alignment of these two spaces.  Like
ABX, this approach requires a distance metric for pairs of stimuli
within each representation space: for the space of neural
representations this distance can be computed as described above for
the ABX metric; for the space of phonological forms a natural choice
is (normalized) edit distance between pairs of phoneme sequences.
\citet{DBLP:journals/corr/abs-2105-05582}
discuss the relation of ABX to the RSA, which can be seen as more
general but less controlled distance-based method of evaluating
representations against each other.

\paragraph{Representation scope}
\citet{chrupala-etal-2020-analyzing} point out that DC is usually
applied at the local scope, i.e.\ to classify a single activation
vector (corresponding to a single or a few frames of the audio), while
the typical usage of RSA involves either pooling over the whole
utterance, or using a distance metric which takes whole utterances
into account such as DTW. They accordingly design a set of experiments
which disentangle scope (local vs global) from the particular metric
(DC vs RSA).  They find out that in some scenarios the presence of
pooling may alter the conclusions: for example for activations
emerging in a visually grounded model of speech, a DC applied to local
activations behaves differently from one applied to globally pooled
activations.

\begin{figure}[htb]
  \centering
  \includegraphics[scale=0.67]{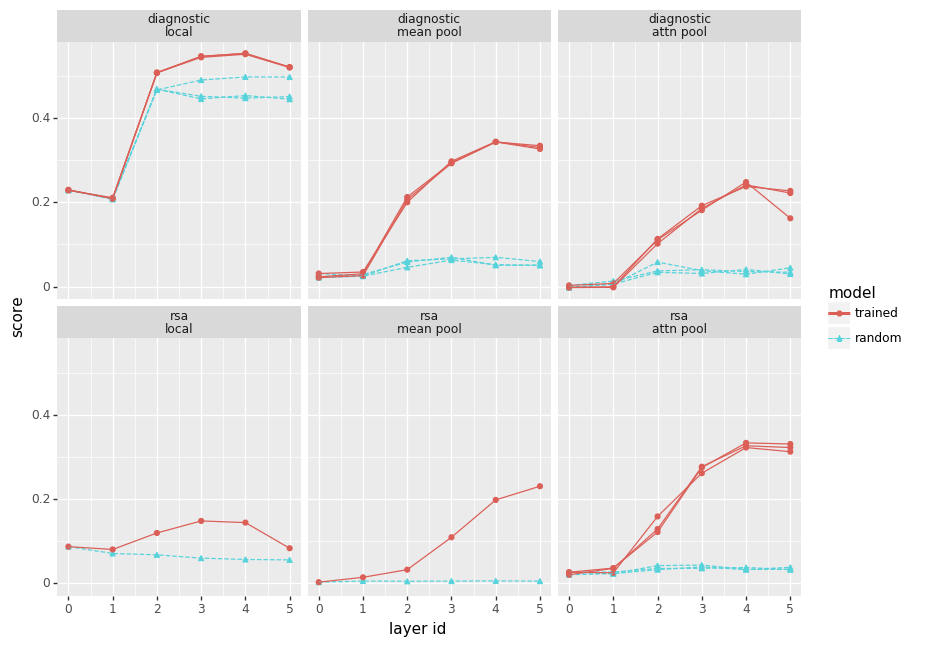}
  \caption{Results of the DC and RSA metrics applied to a visually
    grounded model of spoken language trained on Flickr Audio
    Captions. The score is relative error reduction DC and Pearson's r
    for RSA. {\it Random} refers to activations from a randomly
    initialized but untrained target model. Figure adapted from
    \citet{chrupala-etal-2020-analyzing}.}
  \label{fig:phonological}
\end{figure}

\paragraph{Phoneme encoding}
Overall the application of the above metrics has led to the conclusion
that phonological forms can, to a substantial degree, be decoded from
or correlated with the activation patterns in visually grounded models
of spoken language. There has been less consistency regarding which
layers encode phonology best, and to which extent activations from
untrained models also contain this
information. \Cref{fig:phonological}, reproduced from
\citet{chrupala-etal-2020-analyzing}, summarizes the findings applied
to a GRU-based architecture trained on the Flickr Audio Captions
dataset. This figure shows the scores of the DC and RSA metrics applied
to local (per-frame) activation patters as well as activations pooled
over the whole utterance via mean pooling or an attention-based
pooling mechanism. The scores are shown for activations extracted from
both trained and untrained versions of the target model. The most
striking pattern to note here is that for local activations DC can
extract phoneme information with high accuracy even from activations
from an untrained model, but this is not the case for the pooled activations.

\subsection{Lexicon}
\label{sec:lexicon}
An important concept from linguistics is that of the lexicon, which
can be defined as a mapping relating phonological forms
with their meanings. The lexicon is posited to store those
associations between forms and meanings which are non-compositional,
i.e.\ such that the meaning cannot be straightforwardly predicted from
the form. The content of the lexicon thus consists mostly of small,
basic units such as morphemes and words. Given the lexicon and a set of
rules of composition, the meanings of larger items such as phrases or
sentences can be (largely) computed from the meanings of the constituent
words. We may thus hypothesize that a model trained on the task of
associating the acoustic signal of utterances with correlated visual
features should learn a lexicon-like mapping between word-like
segments in the audio modality, with feature bundles in the visual
modality.
\paragraph{Word presence}
\citet{chrupala-etal-2017-representations} probe for the encoding of
lexicon in a rather simplistic fashion: they test whether the presence
of individual words can be recovered from the utterance representations,
without considering associations with meanings. Their
probe is a Multi-Layer Perceptron classifier applied to the
concatenation of the pooled activation vectors for a layer and a
representation of the target word. For both synthetic and human speech
they saw accuracies substantially above chance, with the strongest
encoding in the middle recurrent layers.

\paragraph{Word activation}
\citet{havard2019word} do not consider lexical mappings either, but
focus on implicit segmentation and word activation, applying the
so-called {\it gating paradigm} (i.e.\ progressive truncation of
stimuli) borrowed from psycholinguistics to probe these phenomena in
an architecture based on \citet{chrupala-etal-2017-representations} as
adapted in \citet{havard2019models}, trained on synthetic speech.
They find that the initial segment of a word (e.g.~/\textipa{dZ9}/ for
{\it giraffe}) activates the corresponding visual concept, where
concept activation is determined by image ranking performance on such
truncated audio inputs. \citet{scholten2020learning} confirm these
findings using a model based on \citet{Merkx2019}, trained on natural
speech from Flickr Audio Captions.

\begin{figure}
  \centering
  \includegraphics[scale=0.15]{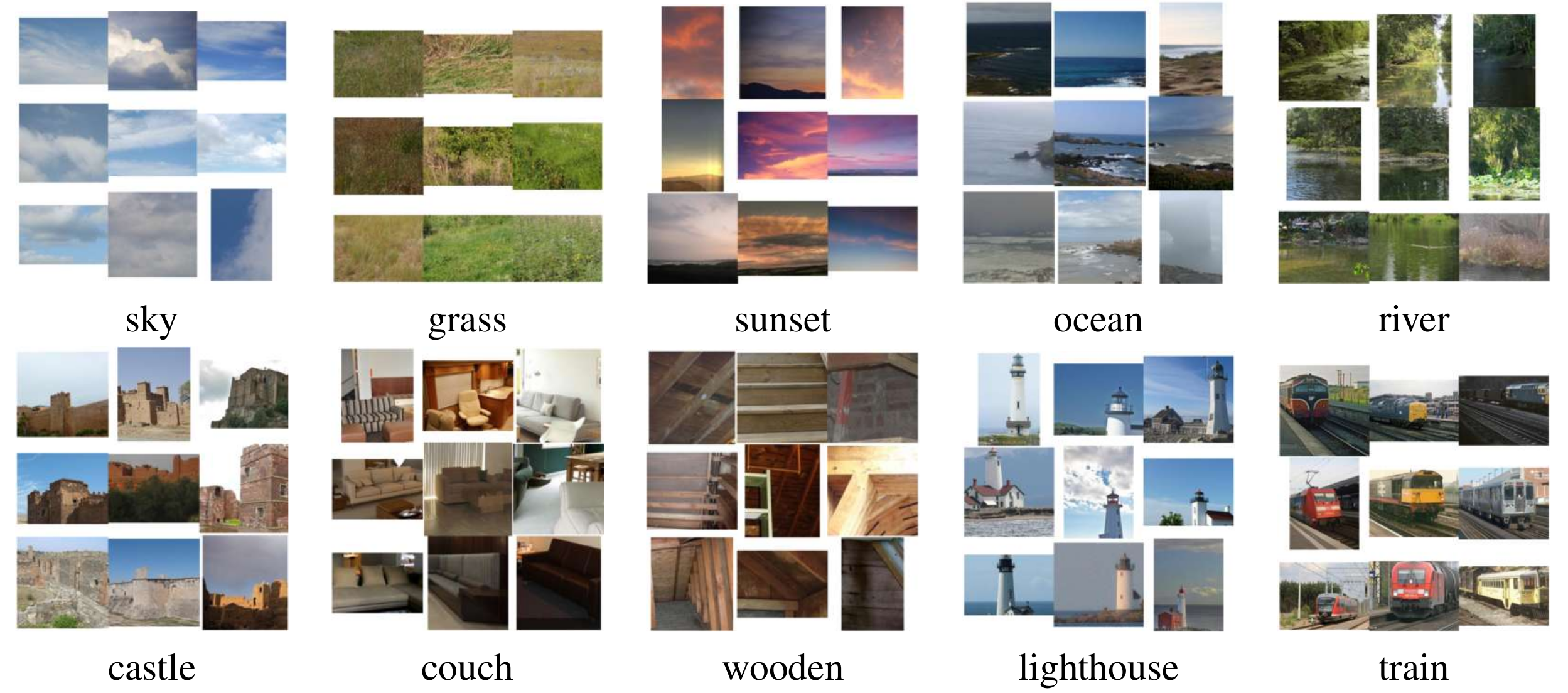}
  \caption{Central image crops from several image clusters, along with
    the label of their most associated acoustic pattern
    cluster. Adapted from \citet{harwath-glass-2017-learning}.}
  \label{fig:lexicon-cluster}
\end{figure}

\paragraph{Word-object associations}
\citet{harwath-glass-2017-learning} take a different approach: they
start with the basic architecture of \citet{harwath2016unsupervised}
and complement it with a dedicated lexicon extraction component. It
works by extracting candidate audio segments as well as candidate
image regions, applying pruning to avoid too much overlap between
candidates. Then the network (which is trained on the Places dataset
in the regular fashion) computes pairwise similarities between each
audio candidate and each visual candidate in the embedding space. The
final step is to apply $K$-means clustering to each modality
separately, and compute an audio-visual affinity between each pair of
clusters by summing over the candidate pairwise similarities for items
in the pair of clusters. As such, this approach is somewhat
reminiscent of association mining techniques used in
\citet{roypentland2002learning} and \citet{yu2005role}. The resulting lexicon is
evaluated in terms of cluster purity metrics as well as
qualitatively. \Cref{fig:lexicon-cluster} shows examples of lexical mappings
obtained.
As discussed in \Cref{sec:timespace}, \citet{harwath2018jointly}
introduces an architecture which directly captures the associations
between audio segments and image regions, via the {\it matchmap}. The
matchmaps for a collection of caption-image pairs can then be
clustered resulting in a audiovisual lexicon. Specifically, Birch
clustering \citep{zhang1996birch} combined with agglomerative
clustering results in a set of 135 lexical items associating words
with objects: this was done with an end-to-end version of the model,
showing that a pre-trained image encoder does not play a role in the
discovery of this lexicon.

\subsection{Semantics}
Compared to phonology and lexicon, there has been somewhat less work
on extrinsic evaluation of utterance embeddings from visually grounded models of
spoken language. On the one hand this is due to the fact that intrinsic
metrics capture at least the visual, task-specific aspects of
semantics already. On the other hand it may be due to the relative
paucity of ground-truth annotations of meaning compared to the easy
availability of phonemic transcriptions and dictionaries.

\paragraph{Correlation with human judgments}
\citet{chrupala-etal-2017-representations} adapt the paradigm of
correlating pairwise sentence similarities in the embedding space with
sentence similarities as elicited from human judges: this is
essentially the same method as RSA discussed in \Cref{sec:phonology},
applied to semantics. They use an existing dataset of human sentence
similarity judgments, SICK \citep{marelli2014sick}, synthesize spoken
versions of the stimuli, and apply the test to a model trained on
Synthetically Spoken COCO. They find a substantial correlation for the
top layers of this model, much above that for mean-pooled MFCC
vectors, but also much lower than for the corresponding models trained
on written captions.

\begin{figure}
  \centering
  \includegraphics[scale=0.3]{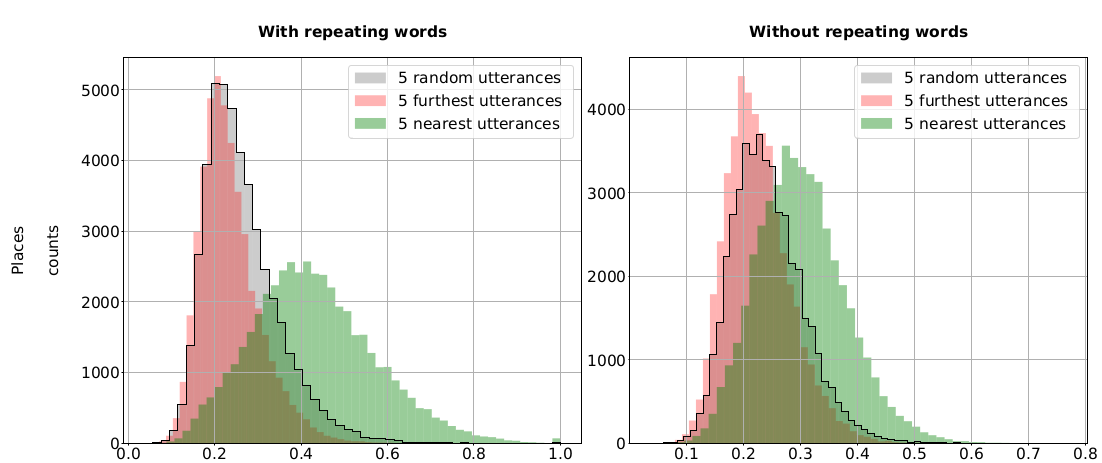}
  \caption{Semantic relatedness scores (SRS) for speech-to-speech
    retrieval for the CNN1 model of \citet{khorrami_2021}. Scores are
    computed between query utterances and the five nearest, the five
    most distant, and five random captions collected for all test
    utterances. Figure adapted from \citet{khorrami_2021}.}
  \label{fig:srs}
\end{figure}

\paragraph{Evaluation based on Word2Vec}
An alternative approach which is easier to apply to human speech was
proposed by \citet{khorrami_2021} and involves using automatically
computed pairwise sentence similarities derived from a text-based
model as the proxy for human similarity judgments of semantic
relatedness. The automatic {\it semantic relatedness score (SRS)} is
based on word-word similarity scores as given by Word2Vec
\citep{mikolov2013efficient} embedding vectors and defined as follows:

\begin{equation}
  \label{eq:srs}
  \mathit{SRS}(r, c) = \mean_{i} \left( \max_{j}
    S_{\mathrm{w2v}}(r_i, c_j) \right)
\end{equation}
where $r$ (reference) and $c$ (candidate) are captions in their
written form (excluding function words) and $S_{\mathrm{w2v}}$ denotes
the Word2Vec similarity score: thus for each word of the utterence $r$
we select the score of the most similar word of utterance $c$ and average
over these scores.  The evaluation consists in computing the SRS
scores between captions corresponding to five closest, five furthest
and five random candidate audio embeddings for each reference
utterance from the test set.  Their results suggest that
visually grounded models of spoken language learn substantial amounts
of the type of distributional semantics that Word2Vec captures:
\Cref{fig:srs} shows the score distributions for the Places dataset.

\paragraph{Encoding of semantics}
Currently our understanding of how and to what extent semantics is
captured by embeddings from visually grounded models of spoken
languages is still limited. It does seem that substantial semantic
information is captured, but it is not clear how task specific it is,
and how much of it is due to network dynamics as opposed to learning.

% LocalWords:  disfluencies embeddings agglomerative dataset MFCC RSA
% LocalWords:  compositional activations convolutional groundable

\section{Conclusion}
We have surveyed the developments in approaches to modeling visual
grounding for spoken language since early 2000s. We have seen that the
first wave of interest in this problem came mostly from the field of
cognitive science and used small-scale datasets and modular
architectures to come up with proof-of-concept systems and answer
questions mainly centered on understanding language learning in human
children. This phase was followed by a second wave of interest largely
driven by the collection of larger-scale datasets, and the application
of increasingly end-to-end neural architectures, with less
human-centered and more AI centered research questions in
mind. Nevertheless, interest in understanding the relation between
representations learned by visually grounded neural models of spoken
language and the central concepts from linguistics has persisted and
has led to many studies focused largely on analytical techniques and
insights to be gained from their application. \Cref{tab:listing} lists
the principal works on visually grounded modeling of spoken language
mentioned in this survey, together with a summary of the data
and modeling approaches proposed, used or analyzed in each paper.

\begin{table}[htb]
  \centering
  \begin{small}
    \begin{tabular}{lllp{3cm}p{3cm}}\toprule
      Publication                     & Grounding & Lang. &  Datasets & Architecture \\\midrule
      \citet{roypentland2002learning} & Images  & en       &  -  & CELL \\
      \citet{yu2005role}              & Images  & en       &  -  &
                                                                   IBM1, EM\\
      \citet{synnaeve2014learning}    & Images  & en       &  Pascal1K, LUCID &
                                                                                        RCNN,
                                                                                        MLP
      \\
      \citet{harwath2015deep}        & Images   & en       &  Flickr & RCNN,
                                                                       CNN \\
      
      \citet{harwath2016unsupervised}& Images   & en      &  Places     & VGG, CNN \\
      \citet{chrupala-etal-2017-representations} & Images & en & Flickr,\newline
                                                                 Synth.~COCO
                                                                      &
                                                                                 VGG,
                                                                                 RHN
      \\
      \citet{alishahi-etal-2017-encoding} & Images & en  &
                                                           Synth.~COCO
                                                                      &
                                                                        VGG,
                                                                        RHN \\
      \citet{harwath-glass-2017-learning} & Images & en  & Places    &
                                                                       VGG,
                                                                       CNN
      \\
      \citet{kamper2017visually}          & Images & en  & Flickr
                                                                      &
                                                                        VGG,
                                                                        CNN,\newline
      keyword spotting\\
      \citet{harwath2018jointly}          & Images & en  & Places
                                                                      &
                                                                        VGG,
                                                                        CNN,\newline
      matchmap\\
      \citet{harwath2018vision}           & Images & en, hi  & Places & VGG,
                                                                        CNN\\
      \citet{boggust2019grounding}        & Video  & en  &
                                                           YouCook2,\newline
                                                           YouTube-8M
                                                                      &
                                                                        VGG,
                                                                        CNN
      \\
      \citet{Merkx2019}                   & Images  & en  & Flickr
                                                                      &     VGG, GRU  \\
      \citet{havard2019models}            & Images  & en, ja &
                                                               Synth.~COCO,\newline
                                                               Synth.~STAIR              
                                                                      &
                                                                        VGG,
                                                                        GRU \\ 
     \citet{chrupala-etal-2020-analyzing} & Images & en  & Flickr
                                                &
                                                  VGG,
                                                  GRU,
                                                                        \newline
                                                                        Transformer
      \\
      \citet{harwath2020learning}         & Images  & en & Places    &    
                                                                 ResNet,
                                                                       VQ \\
      \\
      \citet{rouditchenko2020avlnet}     & Video    & en & Howto100m &
                                                                      ResNet
      \\
      \citet{scholten2020learning}       & Images & en   & Flickr    &
                                                                       ResNet,
                                                                       GRU
      \\
      \citet{ohishi2020trilingual}     & Images  & en, hi, ja &
                                                                Places,
                                                                Jap. Places
                                                                      &
                                                                        VGG,
                                                                        CNN
      \\
      \citet{khorrami_2021}           & Images   & en & Speech-COCO,
                                                        Places &  VGG,
                                                                 CNN,
                                                                 GRU\\
      \citet{DBLP:journals/corr/abs-2105-05582} & Images & en & Flickr &
                                                                ResNet,
                                                                GRU, VQ\\
      \citet{sanabria21_interspeech}           & Images & en & Flickr,
                                                         Places,
                                                         \newline
                                                         Loc.~Narratives &
                                                                      EfficientNet,
                                                                      ResNet50
                                                                       \\
      \citet{monfort2021spokenmoments}   & Video & en & Spoken Moments
                                                                      &
                                                                        ResNet,
                                                                        TSM \\

      \citet{peng2021fastslow} & Video & en & Flickr, Places,\newline
                                              SpokenCOCO             &
                                                                       CNN,
                                                                       Transformer,
                                                                       \newline
                                                                       cross-modal
                                                                       att.\\
      \\
    
      \bottomrule
    \end{tabular}
  \end{small}

  \caption{Main publications on visually-grounded modeling of spoken
    language. The column {\sc Datasets} contains the abbreviated names
    of public datasets used; the column {\sc Architecture} contains keywords
    summarizing the main components of the modeling approach.}
  \label{tab:listing}
\end{table}

\subsection{Summary of main findings}
The field of visually grounded modeling of spoken language is
undergoing rapid developments so it would be premature to draw
definitive conclusions about the optimal way of approaching this
family of problems at this point. However, some preliminary
observations are possible. Neural architectures have been dominant
during the second wave, both in application-oriented and cognitively
motivated work. It is safe to assume that they will continue to play a
role for some time yet. Regarding some specifics: the visual encoders
are almost invariably pre-trained in a supervised fashion -- it is
currently unclear how easy it will be to overcome this
limitation. With regards to the audio encoding architectures, there is
currently no clear winner: convolutional, recurrent and most recently
transformer-based approaches have all been applied successfully.

One interesting pattern has been a substantial degree of convergence
and interaction between application-oriented and cognitively motivated
models: both tend to use similar architectures, training data, and
both also often carry out in-depth quantitative and qualitative
analyses of learning patterns and learned representations. Overall we
consider this a positive trend but future work may need to address
domain-specific issues and limitations in a more focused manner.

\subsection{Challenges for the future}
\label{sec:challenges}
While much progress has been made on developing visually grounded
neural models of spoken language and understanding their behavior,
major challenges remain. The specifics of these problems and their
relative importance of obviously depend on the overall goal we have in
mind when developing a model, but the issues listed below are likely
relevant at least to some extent for both the cognitively motivated
research and for engineering practical applications.

\paragraph{Generalization}
Current approaches to evaluation focus on retrieval tasks closely
related to the objective functions used to train the models, and on
evaluation data from the same distribution as the training data. The
danger with optimizing evaluation scores in this way is that the
research community overfits to these specifics. The scores may keep
going up, but that does not mean that our models generalize beyond
these tasks and especially beyond captioning datasets. Development of
more robust approaches to evaluation is an urgent task.

\paragraph{Ecological validity} From the point of view of applying
these systems to mimic human language acquisition, one issue is the
lack of ecological validity of current datasets.   The language of
captioning datasets is clearly quite specialized, used a small
vocabulary, a restricted range of syntactic structures, and focuses on
particular types of meanings. Images and their captions are guaranteed
to be quite closely related, which is unlike the much less reliable
visual grounding available to babies and children. Likewise, in a
natural setting, there are pervasive confounds between semantics and
non-semantic features of the audio: most obviously the voice and the
image of the person speaking tend to occur together; less obviously
particular people tend to speak about particular objects or
situations. Such confounds are largely avoided in image captioning
dataset by design, as speakers are assigned to images randomly.

If we want a more realistic picture of the capabilities of our models
we need to train them on data which more closely resembles naturally
co-occurring language and visual information. Also from an application
point of view, it would be preferable if systems could learn from
naturally ocurring data instead of relying on datasets which are
expensive to create and only available in a few major languages.

One step in this direction is the use of videos rather than captioned
static images, but current video datasets such as Howto100m are still
very specific and restricted in their domain. Ideally, we would want
to work with data which reflect children's exposure to grounded spoken
language in the real world, via audio and video recordings
\citep[e.g.][]{clerkin2017real}. Currently, the main obstacles to the
use of such datasets is their small size and lack of public
availability due to privacy regulations.

\paragraph{Grounded and non-grounded language} Current approaches have
focused almost exclusively on the purely visually grounded
scenarios. However, both humans and text-based language models
typically acquire as much or more of their knowledge of language
without grounding, purely from language-internal co-occurrence
statistics, via some kind of self-supervision. We foresee that
combining supervision from the visual modality with self-supervision
will become an important development for the field in the near future.
More speculatively, other supervision signals, especially interaction
and grounding in dialog, would also be interesting avenues to explore.

\section*{Acknowledgements}
This survey has been greatly improved thanks to suggestions,
corrections and other feedback from David Harwath, William Havard,
Lieke Gelderloos, Bertrand Higy, Afra Alishahi, and two anonymous
reviewers.  Thank you!

\bibliography{biblio,anthology}

\bibliographystyle{apalike}
\end{document}